\documentclass[10pt,twocolumn,letterpaper]{article}

\usepackage[pagenumbers]{cvpr} % To force page numbers, e.g. for an arXiv version
\usepackage{marvosym}
\usepackage{tabularx}
\usepackage{color}
\usepackage{colortbl}
\usepackage{xcolor}
\usepackage{multirow}
\usepackage{multicol}
\usepackage{wrapfig}
\usepackage{algorithm}
\usepackage{algorithmic}

\definecolor{cvprblue}{rgb}{0.21,0.49,0.74}
\usepackage[pagebackref,breaklinks,colorlinks,citecolor=cvprblue]{hyperref}

\def\paperID{*****} % *** Enter the Paper ID here
\def\confName{CVPR}
\def\confYear{2024}
\renewcommand{\figurename}{Fig}
\title{Follow-Your-Canvas: \\
Higher-Resolution Video Outpainting with Extensive Content Generation}

\author{Qihua Chen$^{1,3}$\footnotemark[2], \quad Yue Ma$^{2}$\footnotemark[2],\quad  Hongfa Wang$^{1,4}$\footnotemark[2], \quad Junkun Yuan$^{1\textrm{\Letter}}$\footnotemark[2]\\
  Wenzhe Zhao$^{1}$, \quad  Qi Tian$^{1}$,\quad  Hongmei Wang$^{1}$, \quad  Shaobo Min$^{1}$, \quad  Qifeng Chen$^{2}$, \quad  Wei Liu$^{1\textrm{\Letter}}$ \\
  $^{1}$Tencent, Hunyuan \quad  $^{2}$HKUST \quad 
  $^{3}$USTC \quad $^{4}$Tsinghua University\\
\url{https://follow-your-canvas.github.io/}
}

\begin{document}
% \maketitle
\renewcommand{\thefootnote}{\fnsymbol{footnote}}

\twocolumn[{
\renewcommand\twocolumn[1][]{#1}%
\maketitle
\begin{center}
    % \captionsetup{type=figure}
    % \vspace{-4.5\baselineskip}
    \includegraphics[width=0.95\textwidth]{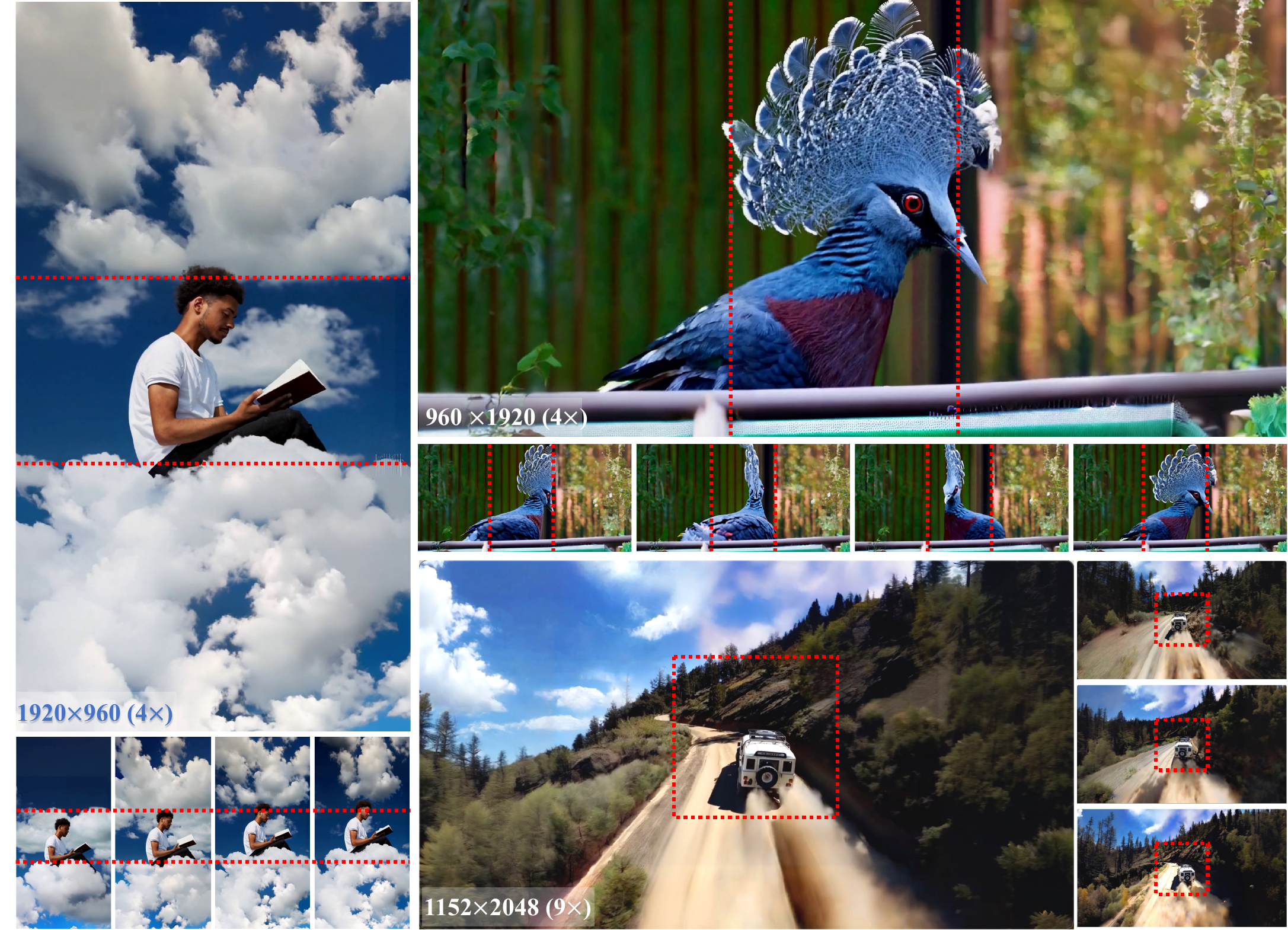}
    % \vspace{-2em}
    \captionof{figure}{
    \textbf{Results of our Follow-Your-Canvas.} The videos (from OpenAI's Sora demo cases) within the red dotted boxes are largely outpainted from 4$\times$ to 9$\times$. Given a video of any size and resolution, Follow-Your-Canvas can generate outpainting results in higher resolution with extensive content, while maintaining consistency of spatial layout, temporal changes, and overall aesthetics. 
    % Given a video of any resolution, the model can product visually smooth and well-structured video clips at any aspect ratio following user instructions. 
    % Text prompt: left - ``a young man reading a book on clouds"; upper right - ``a blue pigeon with blurry plant in the background"; lower right - ``under blue sky with white clouds, a white jeep driving with lots of dust on a narrow road in a forest mountainous area".
%     \textcolor{blue}{([junkunyuan]) Outpainting results of Follow-Your-Canvas on publicly available videos generated by SORA (area within the red dashed box). Although the generated areas are up to 8$\times$, the results are reasonable and in high quality.
\label{fig:result}
    }
\end{center}
}]

\footnotetext[2]{Equal contribution.} \footnotetext[0]{${\textrm{\Letter}}$ Corresponding author.}

\begin{abstract}
This paper explores higher-resolution video outpainting with extensive content generation. We point out common issues faced by existing methods when attempting to largely outpaint videos: the generation of low-quality content and limitations imposed by GPU memory. To address these challenges, we propose a diffusion-based method called \textit{Follow-Your-Canvas}. It builds upon two core designs. First, instead of employing the common practice of ``single-shot'' outpainting, we distribute the task across spatial windows and seamlessly merge them. It allows us to outpaint videos of any size and resolution without being constrained by GPU memory. Second, the source video and its relative positional relation are injected into the generation process of each window. It makes the generated spatial layout within each window harmonize with the source video. Coupling with these two designs enables us to generate higher-resolution outpainting videos with rich content while keeping spatial and temporal consistency. Follow-Your-Canvas excels in large-scale video outpainting, e.g., from $512\times512$ to $1152\times2048$ ($9\times$), while producing high-quality and aesthetically pleasing results. It achieves the best quantitative results across various resolution and scale setups. The code is released on \url{https://github.com/mayuelala/FollowYourCanvas}
\end{abstract}    

\begin{figure*}[t]
    \centering
    \includegraphics[width=.98\textwidth]{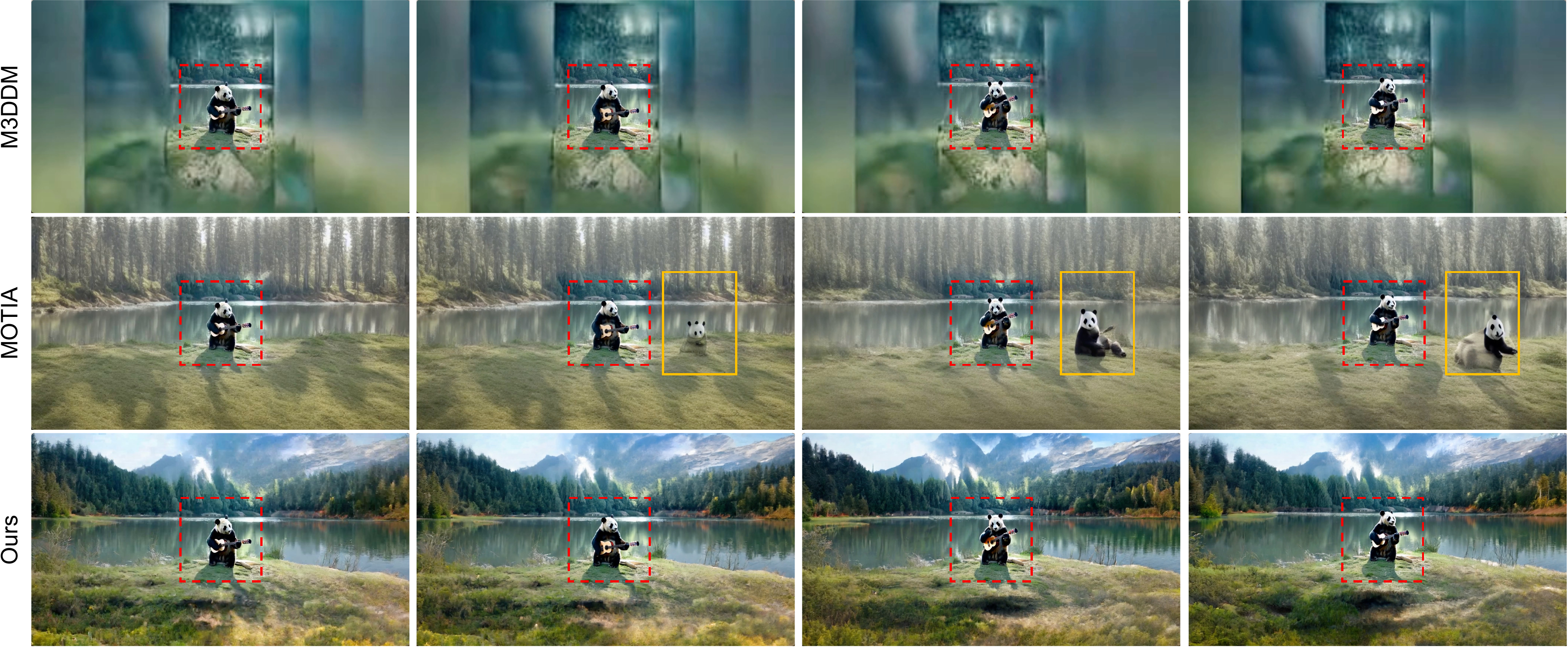}
    \vspace{-0.2cm}
    \caption{\textbf{Results of higher-resolution outpainting with a high content expansion ratio.} The source video (the red dotted box) is outpainted from $512\times512$ to $1152\times2048$ ($9\times$). Existing methods often suffer from blurry content and temporal inconsistencies (yellow boxes). In comparison, our Follow-Your-Canvas method generates well-structured scenes with aesthetically pleasing results.
    } 
    \label{fig:high-res}
\end{figure*}

\begin{figure*}[t]
    \centering
    \includegraphics[width=.98\textwidth]{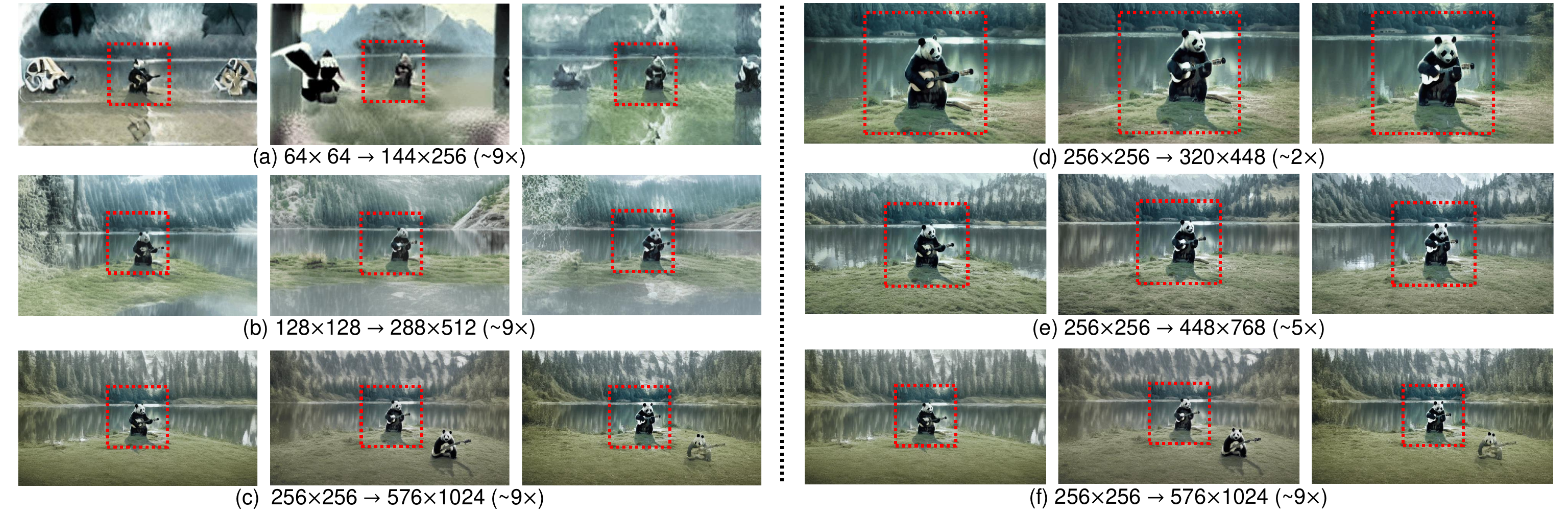}
    \vspace{-0.2cm}
    \caption{
    \textbf{Results of MOTIA with different resolution (a-c) and content expansion ratio (d-f) setups.} Increasing resolution of the source video improves the generation quality, while reducing content expansion ratio improves spatial-temporal consistency.} 
    \label{scale}
\end{figure*}

\begin{figure*}[t]
    \centering
    \includegraphics[width=\textwidth]{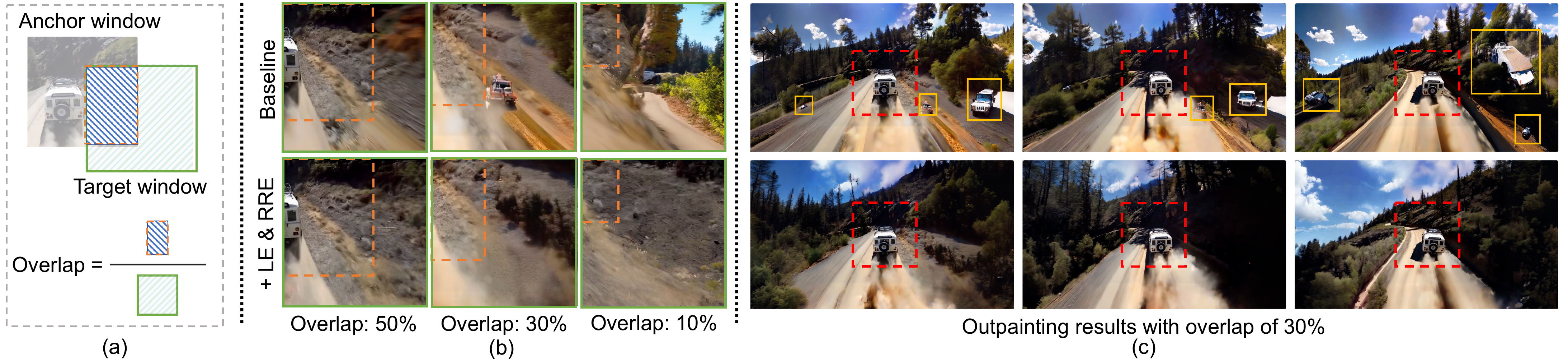}
    \vspace{-0.5cm}
    \caption{
    \textbf{Ablation of layout encoder (LE) \& relative region embedding (RRE).} 
    Under different overlap (a), results within target windows (b) and the final results (c) are presented. 
    The orange dashed line represents the model input for target windows. While the results appear reasonable within windows, they fail to align with the overall layout (see yellow boxes). By incorporating RRE and LE, the model unifies layout of windows with that of the anchor window, improving spatial-temporal consistency. 
    % Zoom in for a detailed view. 
    }
    % } 
    \label{motivation}
\end{figure*}

\section{Introduction}
Video outpainting aims to expand spatial contents of a video beyond its original boundaries to fill a designated canvas region. This task has numerous applications, such as enhancing viewing experience by adjusting aspect ratio of videos to match different users' smartphones~\cite{wang2024your}.

Recently, diffusion models~\cite{ho2020denoising} have emerged as the dominant approach for visual generation, demonstrating exceptional visual synthesis ability by producing appealing results~\cite{rombach2022high}. Meanwhile, several diffusion-based video outpainting methods, such as M3DDM~\cite{fan2023hierarchical} and MOTIA~\cite{wang2024your}, have been proposed. They utilize the source video as a condition and generate the canvas region through step-by-step denoising, showing great performance. However, their results are limited in terms of \textit{resolution}, such as $256\times256$~\cite{fan2023hierarchical} and $512\times1024$~\cite{wang2024your}, or \textit{content expansion ratio}, for example, from $256\times85$ to $256\times256$ ($3\times$)~\cite{fan2023hierarchical} and from $512\times512$ to $512\times1024$ ($2\times$)~\cite{wang2024your}. This raises an intriguing question: \textit{``Is it possible to outpaint a video to higher resolution with a higher content expansion ratio?''} 

This question drives us to evaluate the capability of existing methods in tackling this difficult task. However, we find that they fall short due to limitations in GPU memory. To further explore their potential, we reduce the resolution of the source video through resizing and then resizing it back after outpainting (see details in Section~\ref{sec:experiments}). The results are depicted in Fig~\ref{fig:high-res}. We observe that both M3DDM~\cite{fan2023hierarchical} and MOTIA~\cite{wang2024your} produce low-quality results, e.g., blurry content and temporal inconsistencies. This motivates us to delve deeper into understanding the reasons behind this. We speculate that there are two possible factors contributing to this: (i) the reduced resolution after resizing negatively affects the performance, and (ii) the content expansion ratio is too high to achieve satisfactory results. We conduct experiments with respect to the variations of these factors, see Fig~\ref{scale}. The results demonstrate that both low resolution and a high content expansion ratio significantly reduce generation quality. In other words, achieving high-quality results requires performing outpainting in the \textit{original/high resolution} with a \textit{low content expansion ratio}. 

Based on the analysis above, we propose a diffusion-based method called Follow-Your-Canvas for higher-resolution video outpainting with extensive content generation. We identify that the GPU memory limitations arises from the ``single-shot'' outpainting practice~\cite{fan2023hierarchical, wang2024your}: directly taking the entire video as the input. In contrast, our Follow-Your-Canvas is designed to distribute the task across spatial windows. It kills two birds with one stone. First, it enables us to outpaint any videos to higher resolution with a high content expansion ratio, without being constrained by GPU memory. Second, it simplifies the challenging task by breaking it down into smaller and easier sub-tasks: outpainting each window in the original/high resolution with a low content expansion ratio. Specifically, during the training phase, we randomly sample an anchor window and a target window from the source video, mimicking the ``source video'' and ``outpainting region'' for inference respectively. It helps model learn how to flexibly outpaint with different relative positions and overlaps between the source video and outpainting region. During the inference phase, we outpaint a video by denoising windows that covering the entire video. To accelerate the generation process, we perform window outpainting in parallel on multiple GPUs. After each step of denoising, we seamlessly merge the windows using Gaussian weights~\cite{bar2023multidiffusion} to ensure a smooth transition between them. Due to the fact that videos of any resolution can be covered by a certain number of fixed size windows, while each window is limited within the GPU memory range, our Follow-Your-Canvas method could be applied to situations where the canvas size is very large.

Despite the advantages offered by the spatial window strategy, we observe conflicts between the layout generated within each window and the overall layout of the source video (see Fig~\ref{motivation}). This issue arises due to the fact that the model input for each window is only a portion of the source video. Consequently, while the outpainting results within each window are reasonable, they fail to align with the overall layout, particularly when the overlap is low. To address this challenge, our Follow-Your-Canvas method incorporates the source video and its relative positional relation into the generation process of each window. This ensures that the generated layout harmonizes with the source video. Specifically, we introduce a \textbf{L}ayout \textbf{E}ncoder (LE) module, which takes the source video as input and provides overall layout information to the model through cross-attention. Meanwhile, we incorporate a \textbf{R}elative \textbf{R}egion \textbf{E}mbedding (RRE) into the output of the LE module, which offers information about the relative positional relation. The RRE is calculated based on the offset of the source video to the target window (outpainting region), as well as the size of them. The LE and RRE guide each window to generate outpainting results that conform to the global layout based on its relative position, effectively improving the spatial-temporal consistency. 

Coupling with the strategies of spatial window and layout alignment, our Follow-Your-Canvas excels in large-scale video outpainting. For example, it outpaints videos from $512\times512$ to $1152\times2048$ ($9\times$), while delivering high-quality and aesthetically pleasing results (Fig~\ref{fig:result}). When compared to existing methods, Follow-Your-Canvas produces better results by maintaining spatial-temporal consistency (Fig~\ref{fig:high-res}). Follow-Your-Canvas also achieves the best quantitative results across various resolution and scale setups. For example, it improves FVD from $928.6$ to $735.3$ ($+193.3$) when outpainting from $512\times512$ to $2048\times1152$ ($9\times$) on the DAVIS 2017 dataset.

Our main contributions are summarized as follow:
\begin{itemize}
    \item We emphasize the importance of high resolution and a low content expansion ratio for video outpainting.
    \item Based on the observation, we distribute the task across spatial windows, which not only overcomes GPU memory limitations but also enhances outpainting quality.
    \item To ensure alignment between the generated layout and the source video, we incorporate the source video and its relative positional relation into the generation process.
    \item Our Follow-Your-Canvas demonstrates great outpainting capabilities through both qualitative and quantitative results. 
\end{itemize}

\section{Related Work}
\label{sec:related-work}
% \subsection{Diffusion Models}
\noindent\textbf{Diffusion models}~\cite{ho2020denoising, song2020denoising} are a class of generative models that progressively convert noise into structured data through a learned denoising process. It has garnered significant attention in visual generation~\cite{ramesh2022hierarchical, zhang2023adding, podell2023sdxl, ma2024followyouremoji, ma2023magicstick}. By applying diffusion models in the latent space, LDM~\cite{rombach2022high} has demonstrated the ability to generate high-quality images by utilizing limited computational resources. Meanwhile, many works~\cite{guo2023animatediff, blattmann2023align, ho2022imagen} generate impressive videos by inserting temporal layers into the model structure. 
This has promoted the rapid development of video generation in editing~\cite{ceylan2023pix2video, liu2024video, qi2023fatezero}, controllable generation~\cite{ma2024followyourclick, xue2024followyourposev2, ma2023follow, ma2024followyourclick}, 
% super-resolution~\cite{zhou2024upscale}, 
outpainting~\cite{fan2023hierarchical, wang2024your}, etc. 

\noindent\textbf{Video outpainting} seeks to extend the spatial contents of a video beyond its initial boundaries, allowing it to fill a specific canvas region. Although image outpainting~\cite{zhang2024continuous, yu2024shadow, cheng2022inout} has been extensively studied, video outpainting~\cite{dehan2022complete} still needs to be fully researched. Recently, some diffusion-based approaches have been introduced. M3DDM~\cite{fan2023hierarchical} presents global frame-guided training with a coarse-to-fine inference pipeline to tackle the artifact accumulation issue. Meanwhile, MOTIA~\cite{wang2024your} proposes a test sample-specific fine-tuning strategy to learn the patterns of each sample. Despite their great results, they are limited in terms of resolution such as $256\times256$ and $512\times1024$, or content expansion ratio such as $2\times$ and $3\times$. As these two factors are the core of outpainting, this paper makes the first attempt to study video outpainting with high resolution, \textit{e.g.}, $1152\times2048$, and a high content expansion ratio, \textit{e.g.}, $9\times$.

\begin{figure*}[t]
    \centering
    \vspace{-0.3cm}
    \includegraphics[width=0.95\textwidth]{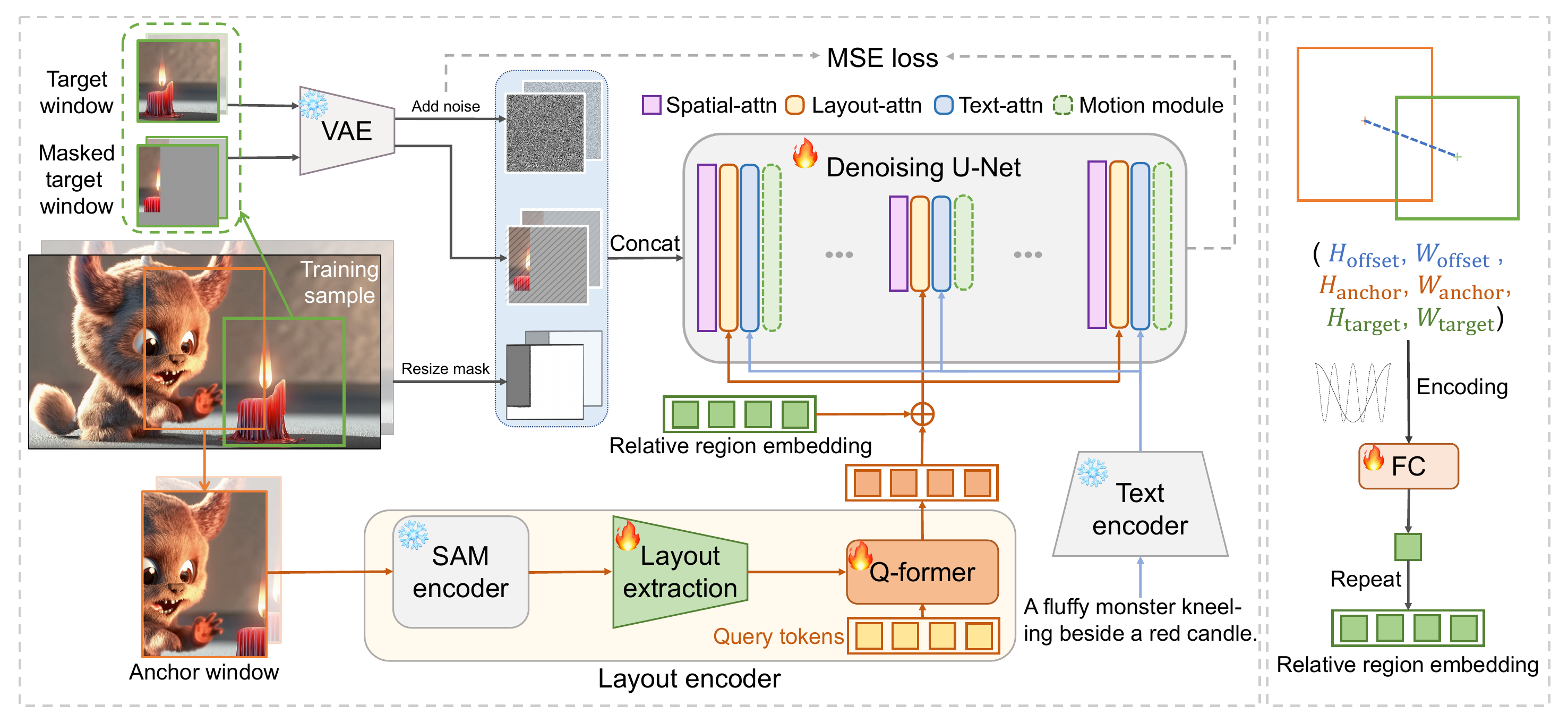}
    \vspace{-0.3cm}
    \caption{
    \textbf{The training phase of Follow-Your-Canvas.} 
    An anchor window and a target window are randomly sampled, mimicking the ``source video'' and ``region to perform outpaint'' for inference respectively.
    % Given a training video sample, it randomly samples an anchor window and a target window, serving as the ``original video'' and the ``region to perform outpainting'' respectively. 
    % The concatenation of the noisy latent representation of the target window, the latent representation of masked target window, and the mask are the model input. 
    The anchor window is injected into the model through a layout encoder, as well as a relative region embedding calculated by the positional relation between the anchor window and the target window, helping the model align the generated layout of the target window with the anchor window. 
    % The proposed layout-aware video outpainting framework. During training, we randomly sample an anchor video at the center of the input video and a target outpainting window. Given the corrupted target latent, the masked latent, and the binary mask, the network predicts the noise added to the target latent. To allow the model to have a perception of the layout in the given anchor video and outpaint a scene that is together consistent with the text prompt, we design a layout encoder and region embedding. 
    % \textcolor{blue}{([junkunyuan]) \textbf{Our Follow-Your-Canvas framework.} Given the corrupted target latent, the masked latent, and the binary mask, Unet learns to outpaint by denoising the target latent.}
    } 
    \label{train}
\end{figure*}

\begin{figure*}[t]
    \centering
    \includegraphics[width=\textwidth]{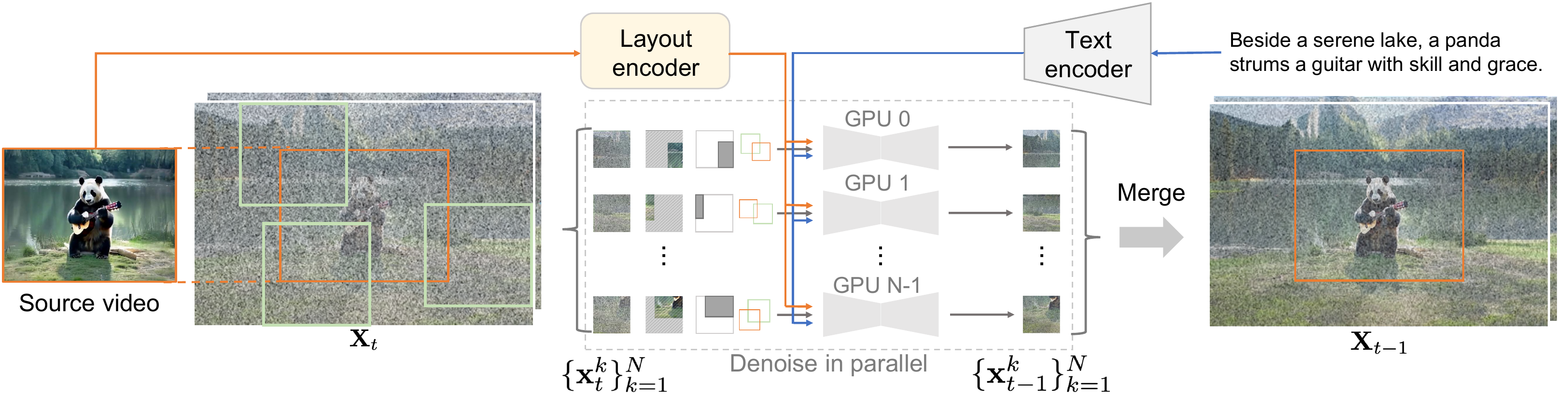}
    \caption{\textbf{The inference phase of Follow-Your-Canvas.} The given source video is covered by $N$ spatial windows. During each denoising step $t$, outpainting is performed within each window in parallel on separate GPUs to accelerate inference. The windows are then merged through Gaussian weights to get the outcome at step $t-1$. Note that these windows may cover layer upon layer, allowing Follow-Your-Canvas to outpaint any videos to a higher resolution without being limited by the GPU memory constraints. 
    % Our high-resolution video outpainting pipeline. In each denoising step, multiple GPUs parallelly denoise $N$ target outpainting windows $\{\mathbf{x}_t^k\}_{k=1}^{N}$ to $\{\mathbf{x}_{t-1}^k\}_{k=1}^{N}$. Then $\{\mathbf{x}_{t-1}^k\}_{k=1}^{N}$ are fused to the entire latent $\mathbf{X}_{t-1}$ following \citet{bar2023multidiffusion}. 
    } 
    \label{infer}
\end{figure*}

\section{Method}
\label{sec:method}

We present Follow-Your-Canvas, a diffusion-based method, which enables higher-resolution video outpainting with extensive content generation. Our approach is built upon two key designs. First, we employ spatial windows to divide the outpainting task into smaller and easier sub-tasks. Second, we introduce a layout encoder module as well as a relative region embedding to align the generated spatial layout.

\subsection{Outpainting by Spatial Windows}
\label{sec:Window-based sampling strategy}
To address the GPU memory limitations, we distribute the outpainting task across spatial windows. 
It allows us to outpaint any videos to higher resolution with a high content expansion ratio without being constrained by GPU memory. Moreover, it simplifies the task by breaking it down into smaller and easier sub-tasks: outpainting each window in its original/high resolution with a low content expansion ratio. 

\noindent\textbf{Training phase.}
Fig~\ref{train} illustrates the training phase of Follow-Your-Canvas.
Given each training video sample, we randomly crop an anchor window and a target window. They serve as the ``source video'' and the ``region to perform outpainting'' respectively, mimicking the source video and the outpainting windows during inference, respectively. The conventional training practice of the latent diffusion model adds noise to the latent representation of the data (the target window) to build the model input and makes the model predict the noise. Here, we concatenate it with conditions: the latent representation of a masked target window and the binary mask. They offer information of the original video and its position. Since the channel of the mask and the latent representations output by the VAE encoder are 1 and 4 respectively, the final model input has 9 channels. We modify the first convolution layer of the denoising UNet to adjust to the channel changes, similar to previous works~\cite{fan2023hierarchical, wang2024your}. 
However, instead of employing a fixed region for outpainting~\cite{fan2023hierarchical, wang2024your}, we use a random sample of the anchor window and the target window.
It helps the model learn to flexibly outpaint with different relative positions and overlaps between the source video and the outpainting region, enabling the sliding window-based inference phase described next. Note that the size of the anchor window, the target window, and their overlap are all variables. See details in experiments. 

\noindent\textbf{Inference phase.}
Fig~\ref{infer} illustrates the inference phase of Follow-Your-Canvas. Given a source video to be outpainted, our Follow-Your-Canvas first determines the number (denoted as $N$) of spatial windows and their positions, which should cover the source video and fill the target region to be outpainted (find more details in experiments). During each denoising step $t$, Follow-Your-Canvas performs outpainting within each window $k$ on noisy data $\mathbf{x}_{t}^k$, where $k\in\{1,...,N\}$. Here, the source video and the window correspond to the anchor window and the target window of the training phase respectively. The denoised outputs in the $N$ windows, i.e., $\{\mathbf{x}_{t-1}^k\}_{k=1}^{N}$, are then merged via Gaussion weights~\cite{bar2023multidiffusion} to get a smooth outcome $\mathbf{x}_{t-1}$. The process is repeated until the final outpainting result $\mathbf{x}_{0}$ is obtained. Importantly, the inference process of each window is independent of the others, allowing us to perform outpainting within each window in parallel on separate GPUs, thereby accelerating the inference. We analyze its efficiency in experiments.

\noindent\textbf{Layout Alignment}
Despite the advantages offered by the spatial window strategy, we observe conflicts between the layout generated within each window and the overall layout of the source video, as shown in Fig~\ref{motivation}. The outpainting results within each window of the ``baseline'', which only applies the spatial window strategy, are reasonable. However, they do not align with the global layout because each window is provided with a view of only a part of the source video. To enable spatial and temporal consistency, we introduce a layout encoder and relative region embedding. They deliver the layout information of the source video and its relative position relation to each window respectively, effectively helping the model generate more stable and consistent outpainting videos (see the results of ``+LE \& RRE'' method in Fig~\ref{motivation}).
% Due to the limitation of paper length, \textit{we leave more specific details about the training recipe, the design of the anchor and target windows, and the inference pipeline in the appendix.}

\noindent\textbf{Layout Encoder (LE).}
Similar to the text encoder that injects the text prompts into the model, we introduce LE to incorporate layout information from the source video, see Fig~\ref{train}. Specifically, LE consists of a SAM encoder~\cite{SAM}, a layout extraction module, and a Q-former~\cite{li2023blip}. Instead of employing the CLIP visual encoder~\cite{radford2021learning} like many previous works~\cite{ye2023ip, xue2024followyourposev2}, we find SAM encoder (ViT-B/16 structure) is more effective to extract visual features by providing finer visual details (see comparisons in experiments). Then, the layout features are extracted by the layout extraction module, including a pseudo-3D convolution layer, two temporal attention layers, and a temporal pooling layer. Inspired by~\citealp{li2023blip}, we employ a Q-former (Querying Transformer) to extract and refine visual representations of the layout information by learnable query tokens. We train the layout extraction module and the Q-former while fixing the SAM encoder. The relative region embedding is added to the output of the LE to provide a positional relation between the anchor window and the target window, introduced next. 

\noindent\textbf{Relative Region Embedding (RRE).} 
RRE provides the positional relation between the anchor window and the target window (see Fig~\ref{train}). We denote the height, width, and center point coordinates of the anchor window as $H_{\text{anchor}}$, $W_{\text{anchor}}$, and $(X_{\text{anchor}}, Y_{\text{anchor}})$ respectively. The target window is defined in the same way. RRE employs sinusoidal position encoding~\cite{zhang2024continuous} to embed the size and relative position relation between the anchor and target windows, i.e., $\{H_{\text{anchor}}, W_{\text{anchor}}, H_{\text{target}}, W_{\text{target}}, H_{\text{offset}}, W_{\text{offset}}\}$, where $H_{\text{offset}}=Y_{\text{target}}-Y_{\text{anchor}}, W_{\text{offset}}=X_{\text{target}}-X_{\text{anchor}}$. The embeddings are then fed to a fully-connected (FC) layer. The output of the FC layer is repeated to match the output of the LE. We incorporate the LE and RRE using a cross-attention layer inserted in each spatial-attention block of the model. 
Due to the limitation of paper length, we leave more details about the design of the model structure in the appendix.

\begin{table*}[t]
 \caption{\textbf{Quantitative comparisons for higher resolution video outpainting with high content expansion ratios.} The resolution of the source video is $512\times 512$. MOTIA is noted by \textcolor[rgb]{0.5,0.5,0.5}{gray} because it is based on test sample-specific fine-tuning. 
 % Since all the baseline methods are performed in low resolution, we reproduce them using their official codes.
 }
  \centering
  \resizebox{1.\linewidth}{!}{
\renewcommand{\arraystretch}{1.1}
\setlength{\tabcolsep}{8pt}
  \begin{tabular}{llcccccc}
  \hline
    Resolution & Method & FVD$\downarrow$ & LPIPS$\downarrow$  & AQ$\uparrow$ & IQ$\uparrow$ & PSNR$\uparrow$ & SSIM$\uparrow$  \\
    \hline
    \multirow{4}{*}{$1280 \times 720$ (720P, $\sim3.5\times$)} 
    & \textcolor[rgb]{0.5,0.5,0.5}{MOTIA~\cite{wang2024your}} & $\textcolor[rgb]{0.5,0.5,0.5}{\underline{473.7}}$ & $\textcolor[rgb]{0.5,0.5,0.5}{\underline{0.418}}$ & $\textcolor[rgb]{0.5,0.5,0.5}{\underline{0.494}}$ & $\textcolor[rgb]{0.5,0.5,0.5}{\underline{0.634}}$ &$\textcolor[rgb]{0.5,0.5,0.5}{\mathbf{15.38}}$ & $\textcolor[rgb]{0.5,0.5,0.5}{0.582}$ \\
    % \cline{2-8}
    & Dehan~\cite{dehan2022complete}& $736.0$ & $0.604$ &  $0.435$ & $0.542$ & $13.95$ & $\underline{0.605}$\\
    % MegicEdit~\cite{} \\
    & M3DDM~\cite{fan2023hierarchical} & $631.3$ & $0.524$ & $0.446$ & $0.556$ &\underline{$15.28$} & $\underline{0.605}$\\
    % & MultiDiffusion~\cite{bar2023multidiffusion}  & $709.5$ & $0.420$ & $\underline{0.498}$ & $\mathbf{0.658}$& $13.64$ & $0.577$\\
    % \hline
    & \textbf{Follow-Your-Canvas (Ours)} & $\mathbf{440.0}$ & $\mathbf{0.390}$ & $\mathbf{0.509}$ & $\mathbf{0.658}$ &$\mathbf{15.38}$ & $\mathbf{0.606}$ \\
    \hline
    \multirow{4}{*}{$1440 \times 810$ (1.5K, $
\sim4.5\times$)} & \textcolor[rgb]{0.5,0.5,0.5}{MOTIA~\cite{wang2024your}} & $\textcolor[rgb]{0.5,0.5,0.5}{\underline{575.9}}$ & $\textcolor[rgb]{0.5,0.5,0.5}{\underline{0.457}}$ & $\textcolor[rgb]{0.5,0.5,0.5}{\underline{0.484}}$ & $\textcolor[rgb]{0.5,0.5,0.5}{\underline{0.648}}$ & $\textcolor[rgb]{0.5,0.5,0.5}{\underline{14.52}}$ & $\textcolor[rgb]{0.5,0.5,0.5}{0.539}$\\ 
    % \cline{2-8}
    & Dehan~\cite{dehan2022complete}& $857.2$  & $0.650$ & $0.415$& $0.543$ & $13.38$ & $\underline{0.553}$\\
    % MegicEdit~\cite{} \\
    & M3DDM~\cite{fan2023hierarchical} & $767.4$ & $0.579$ & $0.447$ & $0.519$ & $14.43$ & $0.542$ \\
    % & MultiDiffusion~\cite{bar2023multidiffusion} & $774.1$ & $0.464$ & $\underline{0.499}$ & $\underline{0.649}$ &$13.44$ & $0.543$ \\
    & \textbf{Follow-Your-Canvas (Ours)} & $\mathbf{486.1}$ & $\mathbf{0.440}$ & $\mathbf{0.505}$ & $\mathbf{0.650}$ & $\mathbf{14.90}$ & $\mathbf{0.559}$\\
    \hline
    \multirow{4}{*}{$2048 \times 1152$ (2K, $9\times$)} 
    & \textcolor[rgb]{0.5,0.5,0.5}{MOTIA~\cite{wang2024your}} &$\textcolor[rgb]{0.5,0.5,0.5}{\underline{928.6}}$ & $\textcolor[rgb]{0.5,0.5,0.5}{\underline{0.587}}$ & $\textcolor[rgb]{0.5,0.5,0.5}{\underline{0.419}}$ &$\textcolor[rgb]{0.5,0.5,0.5}{\underline{0.629}}$ & $\textcolor[rgb]{0.5,0.5,0.5}{\underline{12.45}}$ &$\textcolor[rgb]{0.5,0.5,0.5}{0.524}$\\ 
     % \cline{2-8}
    &Dehan~\cite{dehan2022complete} & $1302.1$ & $0.707$ & $0.394$& $0.607$& $11.40$ & $0.501$\\
    % MegicEdit~\cite{} \\
    & M3DDM~\cite{fan2023hierarchical} &$1181.4$ & $0.691$ & $0.411$ & $0.473$ &$12.43$ & $\underline{0.530}$  \\   
    % & MultiDiffusion~\cite{bar2023multidiffusion}   & $1107.4$ & $0.614$ & $\underline{0.460}$ & $\mathbf{0.658}$ &$11.43$ & $0.509$\\
    & \textbf{Follow-Your-Canvas (Ours)} & $\mathbf{735.3}$ & $\mathbf{0.573}$ & $\mathbf{0.472}$ & $\mathbf{0.657}$ &$\mathbf{12.72}$ & $\mathbf{0.535}$ \\
    \hline
  \end{tabular}}
  \label{high-reso}
\end{table*}

\begin{table}[t]
\renewcommand{\arraystretch}{1.05}
 \caption{\textbf{Quantitative comparisons for low resolution video outpainting.} The source video with different aspect ratios is outpainted to $256\times 256$.
 MOTIA is noted by \textcolor[rgb]{0.5,0.5,0.5}{gray} because it is based on test sample-specific fine-tuning.}
\resizebox{1.\linewidth}{!}{
\renewcommand{\arraystretch}{1.1}
\setlength{\tabcolsep}{5pt}
  \begin{tabular}{@{}l@{}cccc@{}}
  \hline
    method   & PSNR$\uparrow$ & SSIM$\uparrow$ & LPIPS$\downarrow$ & FVD$\downarrow$ \\
    \hline
    \textcolor[rgb]{0.5,0.5,0.5}{MOTIA~\cite{wang2024your}} & $\textcolor[rgb]{0.5,0.5,0.5}{\underline{20.36}}$ & $\textcolor[rgb]{0.5,0.5,0.5}{\mathbf{0.758}}$ & $\textcolor[rgb]{0.5,0.5,0.5}{\underline{0.159}}$ & $\textcolor[rgb]{0.5,0.5,0.5}{\underline{286.3}}$\\
    \hline
    Dehan~\cite{he2022latent} & $17.96$ & $0.627$ & $0.233$ & $363.1$\\
    SDM~\cite{he2022latent} & $20.02$ & $0.708$ & $0.216$ & $334.6$\\
    M3DDM~\cite{fan2023hierarchical} & $20.26$ & $0.708$ & $0.203$ & $300.0$\\
    \textbf{Follow-Your-Canvas (Ours)} & $\mathbf{20.80}$ & $\underline{0.726}$ & $\mathbf{0.160}$ & $\mathbf{242.8}$ \\
    \hline \\
  \end{tabular}}
  \label{low-res}
\end{table}

\section{Experiments}
\label{sec:experiments}
\subsection{Setup}
\noindent\textbf{Dataset.} 
M3DDM~\cite{fan2023hierarchical} use a private dataset with $\sim$5M video samples. 
Here, we employ a random subset ($\sim$1M video samples) of the public Panda-70M dataset~\cite{chen2024panda70m} for training, improving reproducibility of our work. 
% The learning rate is set to $1 \times 10^{-5}$, and the batch size is set to $8$. 
% Eight NVIDIA A800 GPUs are used for both training (50K steps) and inference (40 DDIM steps with classifier-free guidance (cfg) of $7.5$). 

\noindent\textbf{Implementation details.}
Our implementation and model initialization is based on the popular video generation framework of AnimateDiff-V2~\cite{guo2023animatediff}. 
Due to the limitation of paper length, \textit{we leave more specific details about the training recipe, the design of the anchor and target windows, and the inference pipeline in the appendix.}
% The target window size remains fixed at $512\times512$ for both training and inference stages. The minimum overlap ratio between target windows and source video is set to $128$. For more details, please refer to the supplemental material.

\noindent\textbf{Evaluation metrics.}
We first employ metrics of PSNR, SSIM~\cite{wang2004image}, LPIPS~\cite{zhang2018unreasonable}, and FVD~\cite{unterthiner2018towards} by following~\citealp{wang2024your}. 
To evaluate high-resolution video generation, we further utilize aesthetic quality (AQ) and imaging quality (IQ)~\cite{huang2023vbench}, assessing the layout/color harmony and visual distortion (e.g., noise and blur) respectively.
% Following previous works\cite{wang2024your}, the metrics include Peak Signal to Noise Ratio (), Structural Similarity Index Measure (SSIM)~\cite{wang2004image}, Learned Perceptual Image Patch Similarity (LPIPS)~\cite{zhang2018unreasonable}, and Frechet Video Distance (FVD)~\cite{unterthiner2018towards}. 
% Since in our high-resolution video outpainting settings, most content of the result videos are generated and there exist diverse correct answers, 
% We further include the aesthetic quality(AQ) and imaging quality(IQ) proposed by VBench~\cite{huang2023vbench} for video generation quality evaluation without ground-truth. Aesthetic quality evaluates the layout and color richness and harmony, while imaging quality refers to the distortion (\textit{e.g.}, noise, blur) in the generated video clip frames.

\noindent\textbf{Baselines.} We compare our Follow-Your-Canvas with the following baseline methods. (1) ~\citealp{dehan2022complete} use the approach of flow estimation and background prediction. 
% (2)  SDM~\cite{blattmann2023stable} is a baseline model based on stable diffusion, which adopts the first frame and last frame as the condition for per-frame image outpainting. 
(2) M3DDM~\cite{fan2023hierarchical} employs global-frame features to achieve global and long-range information transfer. 
% Since M3DDM only supports outpainting of resolution $256$, we resize the outpainted result to target resolution by bilinear interpolation and paste the given portion of video back to the resized video. 
3) MOTIA~\cite{wang2024your} trains a LoRA~\cite{hu2021lora} to learn patterns of test samples. 
% Since direct inference of MOTIA on our high resolution outpainting setting requires more than $80G$ GPU VRAM, we conduct the experiment on the $4\times$ downsampled resolution and postprocess the result video as aforementioned.  
% 4) MultiDiffusion~\cite{bar2023multidiffusion} also uses the spatial window strategy. 
% The model is trained on Pandas-70M using the same mask strategy as~\cite{fan2023hierarchical}, and the inference follows our parallel window sampling inference strategy. 
We reproduce these baseline methods using their official codes for high-resolution video outpainting and directly cite their results in low-resolution.
% We leave specific implementation details in Appendix.

\begin{figure*}[t]
    \centering
    \includegraphics[width=\textwidth]{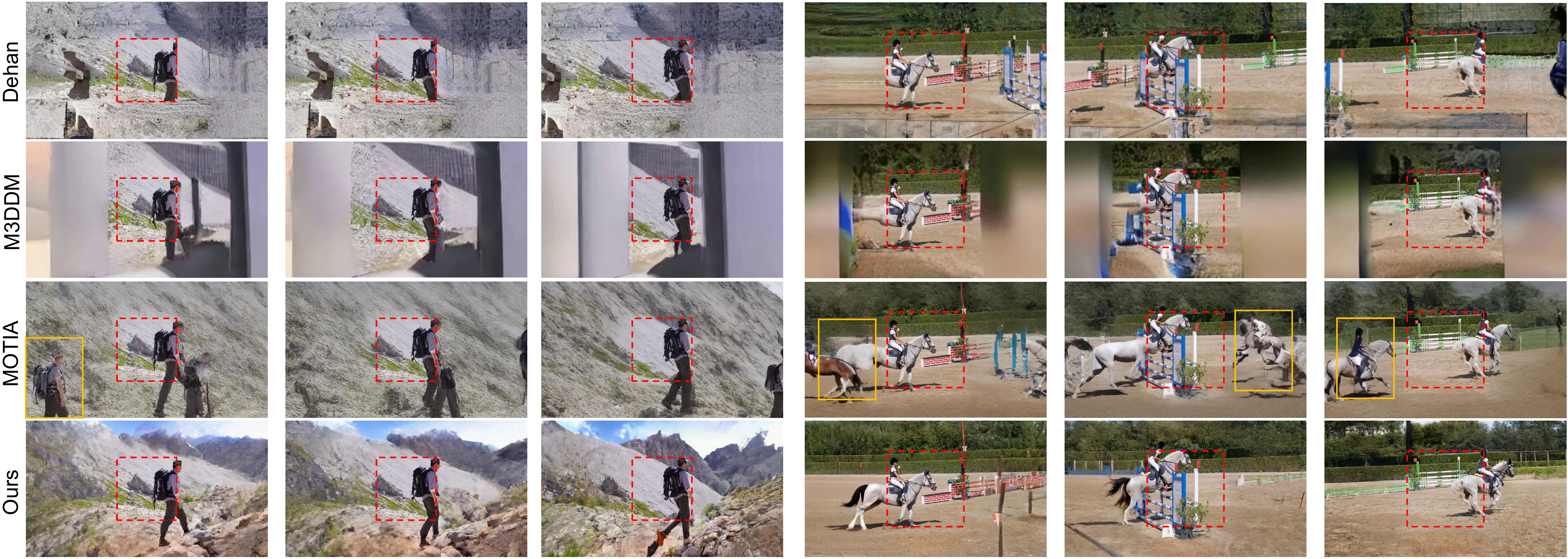}
    \caption{\textbf{Qualitative results.} The source video (the red dotted box) is outpainted from $512\times512$ to $2048\times1152$ (left) or $1440\times810$ (right). Baseline methods suffer from blurry content, and spatial and temporal inconsistencies (yellow boxes).
    % We select four diffusion-based approaches for fair comparison. The red box means input video frames and orange box denotes the inconsistent. It is easy to find that our Follow-Your-Canvas achieves better performance. 
    } 
    \label{davis}
\end{figure*}

\begin{figure*}[t]
    \centering
    \includegraphics[width=\textwidth]{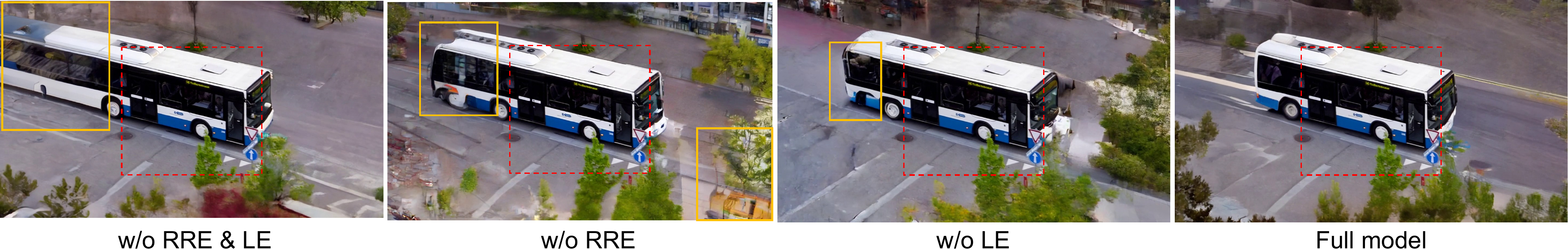}
    \caption{\textbf{Visual results of ablation study.} Layout encoder (LE) and relative region embedding (RRE) effectively guide the generation by providing information of the source video and its positional relation to the outpainting window respectively.
    } 
    \label{ablation}
\end{figure*}

\subsection{Comparisons to Baseline Methods}

\subsubsection{Quantitative results.}
We compare methods in both high and low-resolution settings. 
(1) \textit{High-resolution with large content expansion ratios.}
% We demonstrate our method's superiority in high-resolution video outpainting with high extension ratios. 
% We use the DAVIS 2017 dataset~\cite{perazzi2016benchmark} with full resolution, 
% which has an average size of $1338 \times 2400$. 
% and evaluate them at three target resolutions: $1280\times720$, $1440\times810$, and $2048\times1152$. For each resolution, we resize the original video to the corresponding resolution and crop a $512\times512$ patch in the center as the input video, yielding an expansion ratio of $3\times$, $4\times$, and $9\times$, respectively. 
Table~\ref{high-reso} shows the results. Our Follow-Your-Canvas consistently achieves the best performance for all metrics and outpainting settings. 
Meanwhile, as the resolution and content expansion ratio increase, the performance improvement of many metrics becomes more significant. 
For example, Follow-Your-Canvas improves FVD from 473.7 to 440.0 (+33.7) in 720P ($\sim$3.5$\times$), improves from 575.9 to 486.1 (+89.8) in 1.5K, and improves from 928.6 to 735.3 (+193.3) in 2K. 
Our Follow-Your-Canvas effectively improves performance in the challenging task of high-resolution outpainting with high content expansion ratios.
% , especially for the most challenging task where the original video is expanded by a factor of $9$.  The large improvement in FVD demonstrates the overall structural and motional superiority of the proposed method. Compared to the high-resolution baseline MultiDiffusion, our method produce similar image quality while improves the aesthetic quality of the frames. 
(2) \textit{Conventional settings in low-resolution.}
Following \citealp{fan2023hierarchical} and \citealp{wang2024your}, we also compare results in low-resolution, which outpaint videos to $256\times256$ in the horizontal direction using mask ratio of $0.25$ ($\sim1.3\times$) and $0.66$ ($\sim3\times$) and calculate the average performance. 
Table~\ref{low-res} shows the results.
% perform quantitative comparison of our method and existing works on the DAVIS 2017~\cite{perazzi2016benchmark} dataset with the resolution of $256\times256$, 
% Since our method and MOTIA~\cite{wang2024your} support text prompts, we apply Qwen to extract text prompts from the given portion of the video for fair comparison. 
Our Follow-Your-Canvas still achieves excellent performance under this conventional setting. 
Note that MOTIA~\cite{wang2024your} fine-tunes the model for each test sample which may not be efficient, while our Follow-Your-Canvas method performs zero-shot inference after model training.

% , our proposed method surpass existing finetuning-free methods by a large margin, while achieves comparable results with the one-shot method MOTIA without any sample-specific finetuning. 

\subsubsection{Qualitative results.} 
In Fig.~\ref{davis}, we showcase the qualitative results. It is evident that M3DDM fails to generate meaningful content in the majority of outpainting regions. On the other hand, MOTIA faces difficulties in maintaining spatial and temporal consistencies, which can be attributed to the challenging task of handling high resolution and content expansion ratios. In contrast, our Follow-Your-Canvas successfully generates well-structured visual content. It is because the design of spatial windows that outpaint within each window in its original/high resolution with a low content expansion ratio. Moreover, the layout alignment plays a crucial role in guiding the overall layout of the outpainting results.

\subsection{Ablation Study}
We conduct the ablation study by outpainting the source video from $512\times512$ to $1440\times810$, as shown in Table~\ref{ablation-table}. 
% We study the proposed method without relative rigion embedding (RRE), layout encoder (LE), and layout extraction module respectively. 
We find relative region embedding (RRE), layout encoder (LE), and layout extraction module are all important to achieve the best results. 
Compared to the popular CLIP encoder, we observe that the SAM encoder helps the model to further improve outpainting results. 
Visual results are shown in Fig~\ref{ablation}.

\begin{table}[ht]
\renewcommand{\arraystretch}{1.05}
 \caption{\textbf{Ablation study.} 
 % 1.5K, $\sim4.5\times$ outpainting.
 }
\resizebox{1.\linewidth}{!}{
\renewcommand{\arraystretch}{1.2}
\setlength{\tabcolsep}{5pt}
  \begin{tabular}{@{}l@{}cccc@{}}
  \hline
    Method   & PSNR$\uparrow$ & SSIM$\uparrow$ & LPIPS$\downarrow$ & FVD $\downarrow$ \\
    \hline
    w/o LE \& RRE & $13.44$ & $0.527$ & $0.464$ & $774.1$ \\
    w/o LE & $14.02$ & $0.542$ & $0.450$ & $512.2$\\
    w/o RRE & $13.63$ & $0.532$ & $0.458$ & $670.3$\\
    w/o layout extraction & $13.77$ & $0.535$ & $0.456$ & $550.2$\\
    w/ CLIP image encoder & $14.56$ & $0.553$& $0.441$ & $506.8$\\
    \textbf{Follow-Your-Canvas (ours)} & $\mathbf{14.90}$ & $\mathbf{0.559}$ & $\mathbf{0.440}$ & $\mathbf{486.1}$ \\
    \hline \\
  \end{tabular}}
  \label{ablation-table}
\end{table}

\begin{table}[t]
\renewcommand{\arraystretch}{1.05}
 \caption{\textbf{Run time (minutes).} Parallel inference for outpainting a video of $512\times512$ resolution with 64 frames.}
\resizebox{1.\linewidth}{!}{
\renewcommand{\arraystretch}{1.1}
\setlength{\tabcolsep}{7pt}
  \begin{tabular}{l|cccc}
  \hline
    Resolution & 1 GPU & 2 GPUs & 4 GPUs & 8 GPUs \\
    \hline
    $1280\times720$ & $25.2$ & $14.8$ & $7.8$ & $4.3$ \\
    $1440\times810$ & $58.3$ & $33.5$ & $18.2$ & $11.5$ \\
    $2048\times1152$ & $85.8$ & $51.9$ & $28.9$ & $16.2$ \\
    \hline
  \end{tabular}}
  \label{run-time}
\end{table}
\section{Conclusion}
\label{sec:conclusion}
Largely expanding an image/video is the core of the outpainting task. In this study, we take the first step towards exploring higher-resolution video outpainting with high content expansion ratios. We achieve this by introducing the spatial window strategy combined with the design of layout alignment. Our Follow-Your-Canvas method allows for large-scale video outpainting,
% with aesthetically pleasing results
e.g., from $512\times512$ to $1152\times2048$ (9$\times$). We hope our work can pave the way for further progress in this promising direction and push this frontier.

\noindent\textbf{Limitations.}Although Follow-Your-Canvas has achieved great outpainting performance, it may have a longer inference time due to the spatial window strategy, as shown in Table~\ref{run-time}. To reduce time consumption, we suggest users utilize multiple GPUs in parallel. Besides, we encourage further research to investigate techniques for improving inference speed.

{
    \small
    \bibliographystyle{ieeenat_fullname}
    \bibliography{main}
}

\setcounter{secnumdepth}{2} %May be changed to 1 or 2 if section numbers are desired.
\renewcommand{\figurename}{Fig}
% The file aaai25.sty is the style file for AAAI Press

% Title

% Your title must be in mixed case, not sentence case.
% That means all verbs (including short verbs like be, is, using,and go),
% nouns, adverbs, adjectives should be capitalized, including both words in hyphenated terms, while
% articles, conjunctions, and prepositions are lower case unless they
% directly follow a colon or long dash
% \title{Follow-Your-Canvas: \\ 
% Super High Resolution Video Outpainting with Diffusion models}
% \title{Infinite-Canvas: Video Outpainting with Large Content Expansion}
% \title{Infinite-Canvas: Higher Resolution Video Outpainting with Extensive Content Generation: Supplementary Material}
% boundless, OmniExpand, UltraExpand, VastOutpaint

% InfiniteCanvas: boundless high resolution video outpainting with diffusion models

% Boundless-Canvas, Boundless-Brush
% REMOVE THIS: bibentry
% This is only needed to show inline citations in the guidelines document. You should not need it and can safely delete it.
% \usepackage{bibentry}
% END REMOVE bibentry

% \begin{document}

% \maketitle
\clearpage

\section{More Implementation details}
\subsection{Benchmark}
The quantitative metric evaluation of our method is based on the DAVIS~\cite{perazzi2016benchmark} dataset. The DAVIS (Densely Annotated VIdeo Segmentation) dataset is pivotal for video object segmentation research. Following~\citealp{fan2023hierarchical} and~\citealp{wang2024your}, we use the DAVIS 2017 TrainVal subset, which contains $90$ videos for evaluating the outpainting performance. For the task of high-resolution video outpainting, we use the DAVIS 2017 dataset with full resolution, which has an average resolution of $1338 \times 2400$. For the task of low-resolution video outpainting, we use the $480$p version of the DAVIS dataset following~\citealp{fan2023hierarchical}. 

% Following previous works\cite{wang2024your}, 
We employ the popular metrics including Peak Signal to Noise Ratio (\textit{PSNR}), Structural Similarity Index Measure (\textit{SSIM})~\cite{wang2004image}, Learned Perceptual Image Patch Similarity (\textit{LPIPS})~\cite{zhang2018unreasonable}, and Frechet Video Distance (\textit{FVD})~\cite{unterthiner2018towards}, similar to previous works~\cite{fan2023hierarchical, wang2024your}. 
% Since in our high-resolution video outpainting settings, most content of the result videos are generated and there exist diverse correct answers, 
We further include metrics of aesthetic quality (\textit{AQ}) and imaging quality (\textit{IQ}) from VBench~\cite{huang2023vbench} for video generation quality evaluation (without ground-truth). Specifically, AQ evaluates the layout/color richness and harmony, while IQ assesses the visual distortion such as noise and blur.
 \subsection{Baseline Methods}
We reproduce the baseline methods using their official codes for high-resolution video outpainting and directly cite their results in low-resolution.
Specifically, since M3DDM only supports $256$-resolution outpainting, we resize the source video to perform outpainting, and resize the outpainting video to the target resolution by bilinear interpolation. 
We conduct other methods in the same way if they are constrained by the GPU memory. 
Although it is not fair enough for comparison, our Infinite-Canvas achieves the best results for both the high-resolution and the low resolution tasks.

\subsection{Training of Infinite-Canvas}
The main training recipe of Infinite-Canvas is given below.
The learning rate is set to $1 \times 10^{-5}$, and the batch size is set to $8$. 
Eight NVIDIA A800 GPUs are used for both training (50K steps) and inference (40 DDIM steps with classifier-free guidance (cfg) of $7.5$). The target window size remains fixed at $512\times512$, and the anchor window size, i.e., $H_{\text{anchor}}$ and $ W_{\text{anchor}}$, is sampled from a uniform distribution $\mathrm{U}(512, 1536)$. Note that the anchor window size is the same as the size of the given source video for inference. The minimum overlap between the target window and the source video is set to $128$. Meanwhile, the minimum overlap between the adjacent target windows are also set to $128$.

\subsection{Inference of Infinite-Canvas}
\begin{figure*}[h]
    \centering
    \includegraphics[width=\textwidth]{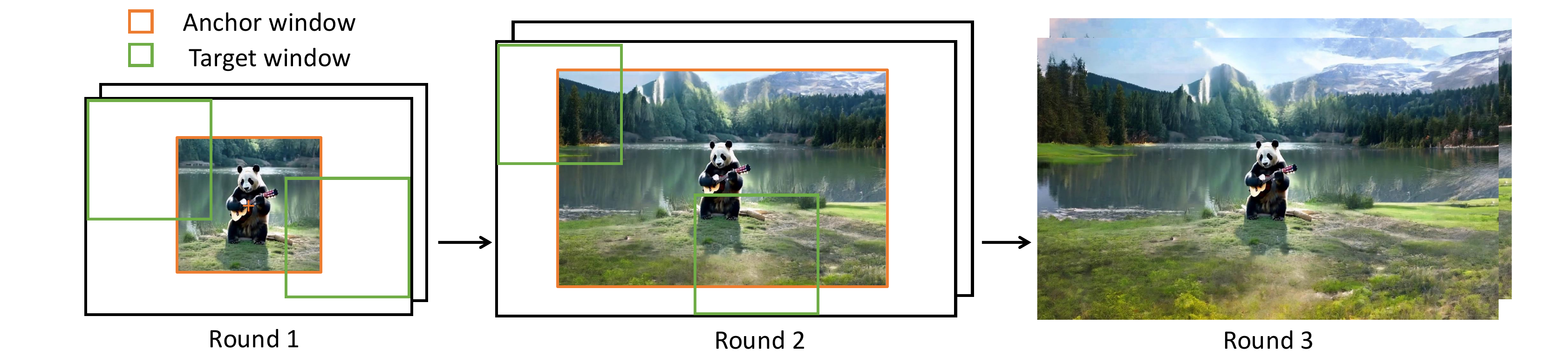}
    \caption{\textbf{Inference pipeline of Infinite-Canvas for high-resolution source videos.} Infinite-Canvas outpaints the high-resolution source videos round by round. Note that the actual target windows should be dense enough to cover the outpainting area. The pipeline is implemented in parallel on separate GPUs to improve efficiency.
    } 
    \label{infer-supp}
\end{figure*}

After training the model using the spatial window strategy, we can outpaint a video from any resolution to any target resolution by dividing the outpainting area into multiple windows and blending the denoising results. 
% By considering the source video and the positional relation between the source video and target windows, 
Specifically, we partition the outpainting region into spatial windows and perform outpainting in multiple rounds, as shown in Figure~\ref{infer-supp}. In the first round, the source video acts as the ``anchor window'', while subsequent rounds utilize the outpainting results from the previous round as the anchor window. This process is repeated until the designated canvas is filled. See the inference pipeline of Infinite-Canvas in Algorithm~\ref{alg}.

\begin{algorithm*}
\caption{Inference pipeline of Infinite-Canvas
% The proposed high-resolution video outpainting pipeline. Note that we bypass the VAE encoding and decoding for simplicity.
}
\begin{algorithmic}[1]
\label{alg}
\footnotesize
\REQUIRE $V_{\text{source}}$: a source video of size $H_{\text{source}} \times W_{\text{source}}$, $\mathcal{\theta}$: the Infinite-Canvas model, $H_{\text{target}} \times W_{\text{target}}$: target size, $T$: total denoising steps, $\{\text{GPU}_0, \text{GPU}_1, ..., \text{GPU}_{N-1}\}$: $N$ available GPUs \\
% \# Get the canvas size for each round
\STATE $N, \{H_0...H_{N}\}, \{W_0...W_{N}\}\leftarrow\texttt{split\_round}(H_{\text{original}}, W_{\text{original}}, H_{\text{target}}, W_{\text{target}})$
\STATE $V_{\text{anchor}} \leftarrow V_{\text{source}}$
\FOR{$i = 1$ to $N$}
    \STATE $V^0 \leftarrow \texttt{initialize\_noise}(H_i, W_i)$
    \FOR{$t = 0$ to $T-1$}
    \STATE ${V_0^t, ..., V_K^t} \leftarrow \texttt{split\_windows}(V_t, H_i, W_i, H_{\text{target}}, W_{\text{target}})$
        \FOR{$\text{GPU} = 0$ to $N-1$}
            % \STATE \text{Each GPU,} $GPU_m$, $m \in \{0, ..., N-1\}$, \text{performs:}
                \STATE \hspace{\algorithmicindent} \text{get} $k \in \{0, ..., K\}$
                \STATE \hspace{\algorithmicindent} $\text{RRE}_k \leftarrow \texttt{get\_relative\_region\_embedding}(k)$
                \STATE \hspace{\algorithmicindent}  $\hat{V_k^t} \leftarrow \mathcal{\theta}(V_{\text{anchor}}, V_k^t, \text{RRE}_k, t)$ on $\text{GPU}_m$
            \STATE $V^{t+1} \leftarrow \texttt{blend\_windows}(V_0^t, ..., V_K^t)$
        \ENDFOR
    \ENDFOR
    \STATE $V_{\text{anchor}} \leftarrow V^{T}$
\ENDFOR
\STATE $V_{\text{outpaint}} \leftarrow V_{\text{anchor}}$
\RETURN $V_{\text{outpaint}}$

\end{algorithmic}
\end{algorithm*}

\section{Preliminaries}
% In this section, we introduce the preliminaries of video latent diffusion models that our framework is based on.

\subsection{Video Latent Diffusion Models}
Diffusion models~\cite{ho2020denoising, li2024hunyuan, yuan2024hap} consist of two processes: a diffusion/forward process that gradually adds Gaussian noise to the clean data using a fixed Markov chain with $T$ steps, and a denoising/reverse process where the trained model generates samples from Gaussian noise. 
Building upon the diffusion model, the latent diffusion model (LDM)~\cite{rombach2022high} performs both the diffusion and denoising processes in a latent space to achieve efficient learning. Specifically, LDM encodes the raw pixels $\mathbf{x}$ into a latent space using a VAE~\cite{kingma2013auto} encoder $\varepsilon$, that is, $\mathbf{z} = \varepsilon(\mathbf{x})$. Meanwhile, the original pixels $\mathbf{x}$ can be approximately reconstructed from the latent representation $\mathbf{z}$ using a VAE decoder $\mathcal{D}$, that is, $\mathcal{D}(\mathbf{z}) \approx \mathbf{x}$. 

In this work, we build our Infinite-Canvas model upon the video latent diffusion model~\cite{guo2023animatediff} for video generation. It inflates the 2D layers of LDM into pseudo-3D layers, incorporating temporal information. It also introduces a temporal motion module to each spatial module in LDM, enabling the model to generate smooth and stable videos.
In the latent space, a Unet~\cite{ronneberger2015u} $\varepsilon_{\theta}$ estimates the added noise  guided by the objective:
\begin{equation}
\label{eq:diffusion-obj}
    \min_{\theta}E_{z_{0},\varepsilon\sim N(0,I),t\sim\text{U}(1,T)}\left\|\varepsilon-\varepsilon_{\theta}\left(z_{t},t,C\right)\right\|_{2}^{2},
\end{equation}
where $C$ is the condition and $z_t$ is a noisy sample of $z_0$ at timestep $t$. 
During inference, given input noise $z_T$ sampled from a Gaussian distribution, network $\varepsilon_{\theta}$ denoises $z_t$ step-by-step and decodes the final latent representation by $\mathcal{D}$. 

\begin{figure}
    \centering
    \includegraphics[width=\linewidth]{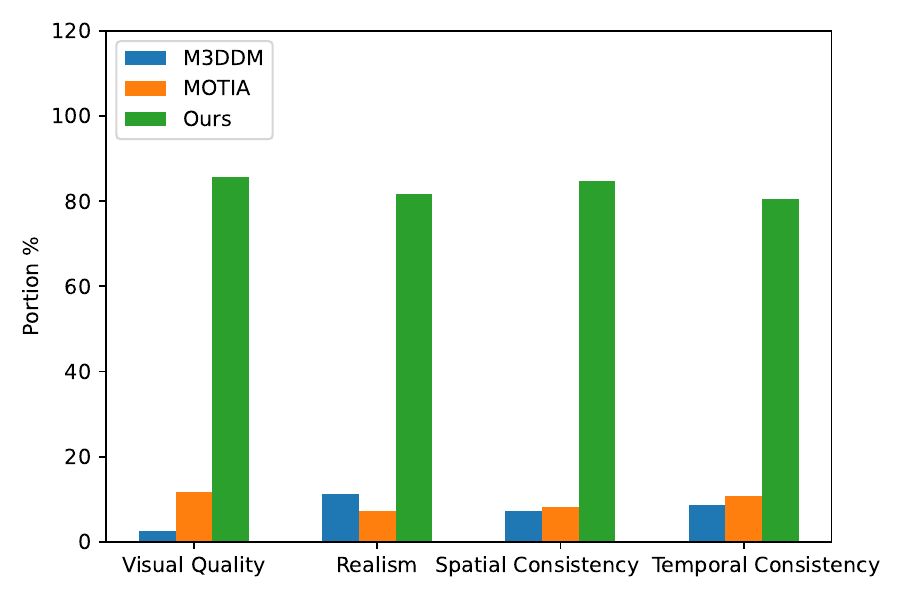}
    \caption{\textbf{User Study.} 30 volunteers are invited to blindly select the best result based on different dimensions.
    } 
    \label{user}
\end{figure}

\subsection{Diffusion-based Video Outpainting}

Video outpainting aims to generate the surrounding regions of a given source video, which can be considered as a conditional video generation task. Its key objective is to make the generated video not only exhibit well-structured spatial layout but also preserves temporal consistency. Following~\citealp{fan2023hierarchical, wang2024your}, we denote the original pixels as $\mathbf{x}$, a 0-1 binary mask as $\mathbf{m}$, the known region as $\mathbf{x}^\text{known}=(1-\mathbf{m})\odot\mathbf{x}$, and the unknown region as $\mathbf{x}^\text{unknown}=\mathbf{m}\odot\mathbf{x}$, where $\odot$ represents Hadamard product. We concatenate the noisy latent representation of the source video, i.e., $\mathbf{z}_T$, with its context as a condition, including the latent representation of the masked video $\mathbf{z}^\text{known}_0$ and the mask $\mathbf{m}$ after resizing. Model parameters $\theta$ is trained by
\begin{equation}
\label{eq:diffusion-obj}
    \min_{\theta}\mathbb{E}_{\mathbf{z},\epsilon\sim \mathcal{N}(0,I),t\sim\text{U}(1,T)}\left\|\epsilon-\epsilon_{\theta}\left(\mathbf{z}_{t},t,C\right)\right\|_{2}^{2},
\end{equation}
where the condition is: $C = \left\{\mathbf{z}^{\text{known}}, \mathbf{m}, e_{\text{text}} \right\}$, and $e_{\text{text}}$ represents the text embedding extracted from a text prompt.

\begin{figure*}[t]
    \centering
    \includegraphics[width=\linewidth]{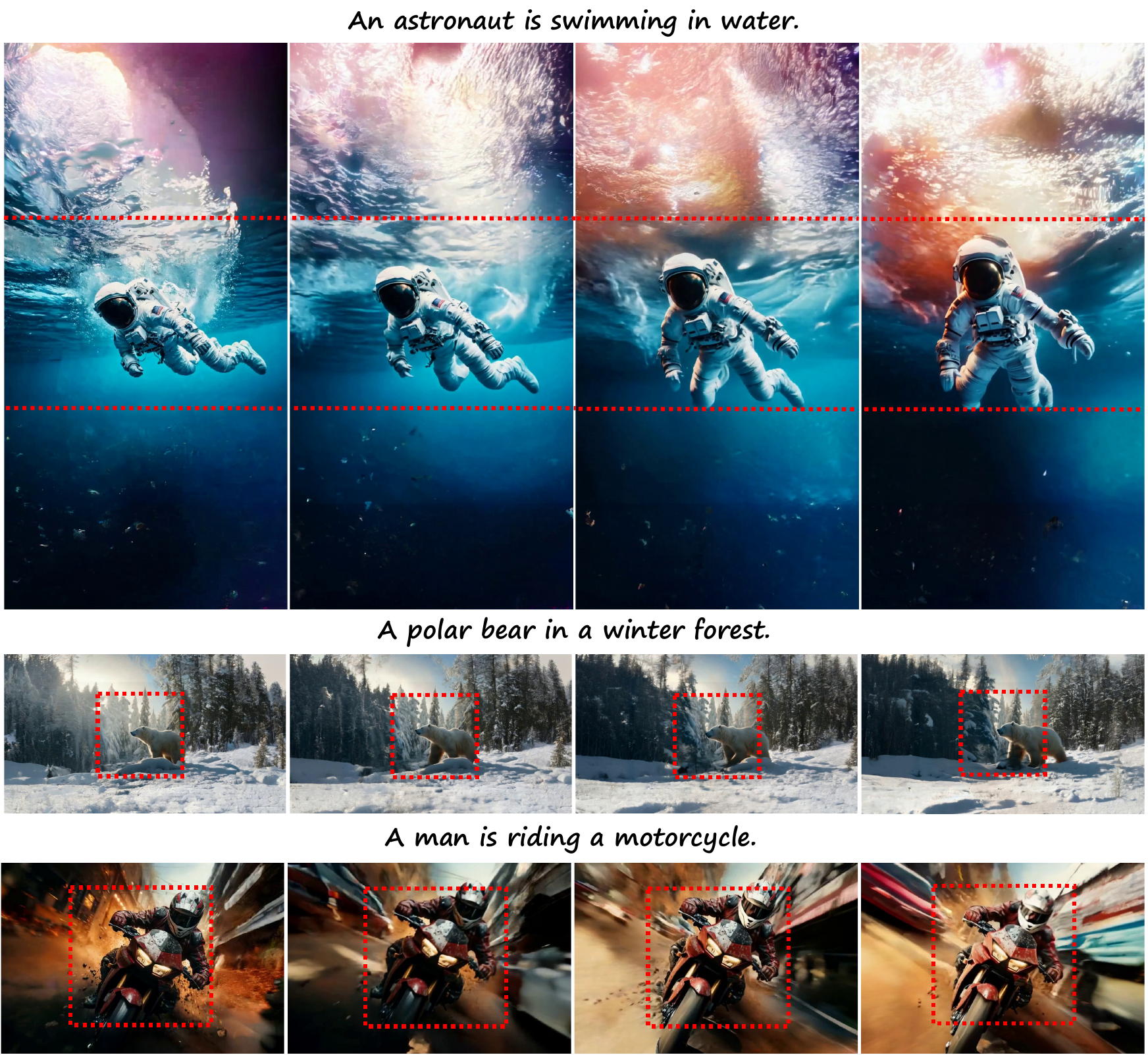}
    \caption{\textbf{More results of Infinite-Canvas.} Infinite-Canvas outpaints source videos with different resolution and styles.
    % We provide various resolution outpainting results, such as $1920 \times 960 (4 \times)$, $960 \times 1920 (4 \times)$, $1152 \times 2048 (9 \times)$.
    } 
    \label{More_visual_outpainting}
\end{figure*}

\section{Additional Results}
\subsection{User Study}
We further conduct a user study comparing our method with MOTIA and M3DDM. We use the DAVIS dataset to outpaint the source video from $512\times512$ to $1440\times810$ resolution. We collect preferences from 30 volunteers, who evaluate 50 randomly selected sets of results based on visual quality (including clarity, color fidelity, and texture detail), realism (whether the overall outpainted scene is harmonious), spatial consistency, and temporal consistency. As shown in Fig.~\ref{user}, the results from our Infinite-Canvas method is overwhelmingly preferred over the other baseline methods.

\begin{figure*}[t]
    \centering
    \includegraphics[width=\linewidth]{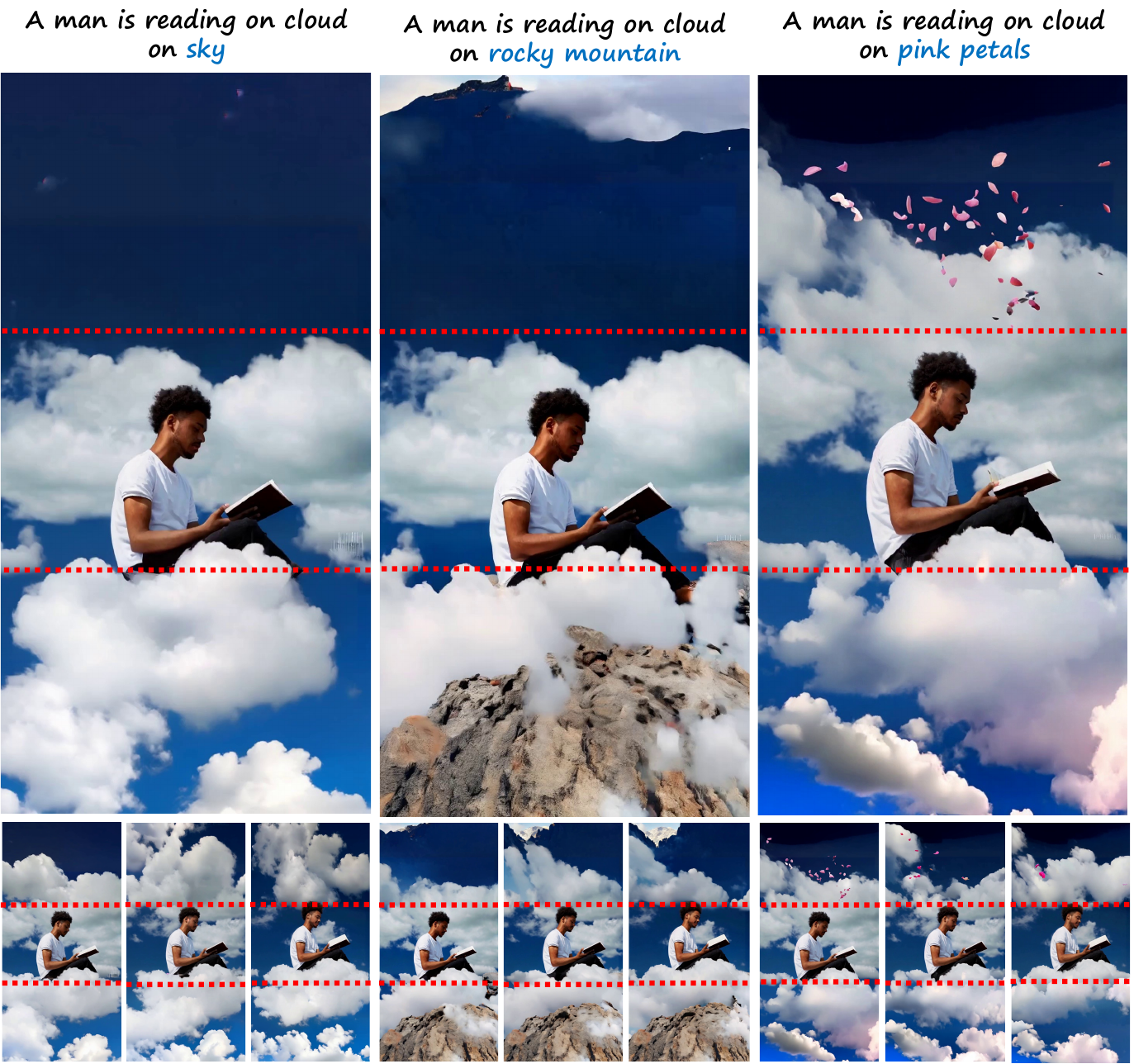}
    \caption{ \textbf{The qualitative results of prompt-following.} We outpaint a source video with various text prompts. It is intriguing to find that our Infinite-Canvas enables one to effectively control the generated contents of outpainting region.
    } 
    \label{various_prompt}
\end{figure*}

\subsection{Prompt-Following Results}
% \subsubsection{Results with different prompts}
\label{prompt}
Since our Infinite-Canvas is based on Animatediff with a text encoder, it naturally supports controlling the generated content using text prompts. We provide three different prompts for outpainting a source video, as shown in Fig.~\ref{various_prompt}. It is interesting to find that our Infinite-Canvas enables one to control the outpainting contents using different text prompts.


\end{document}